\documentclass[runningheads]{llncs}
\usepackage{graphicx}
\usepackage{soul,color}
\usepackage{amsfonts}
\usepackage{acronym}
\usepackage{relsize}
\usepackage{orcidlink}
\usepackage{changepage,tabularx,booktabs}
\acrodef{PH}{Persistent Homology}
\acrodef{DoG}{Difference of Gaussians}
\acrodef{TTK}{Topology ToolKit}
\acrodef{TDA}{Topological Data Analysis}

\begin{document}
\title{Alzheimer Disease Detection from Raman Spectroscopy of the Cerebrospinal Fluid via Topological Machine Learning}
\titlerunning{Alzheimer Detection  via Topological Machine Learning}
%
%

\author{Francesco Conti\inst{1,2}\orcidlink{0000-0001-7393-4267},
Martina Banchelli\inst{3}\orcidlink{0000-0001-5348-0552}, Valentina Bessi\inst{4}\orcidlink{0000-0002-6176-3584}, Cristina Cecchi\inst{5}\orcidlink{0000-0001-8387-7737}, Fabrizio Chiti\inst{5}\orcidlink{0000-0002-1330-1289}, Sara Colantonio\inst{1}\orcidlink{0000-0003-2022-0804}, Cristiano D'Andrea \inst{3}\orcidlink{0000-0001-5807-3067}, Marella de Angelis\inst{3}, Davide Moroni\inst{1}\orcidlink{0000-0002-5175-5126}, Benedetta Nacmias\inst{4,6}\orcidlink{0000-0001-9338-9040}, Maria Antonietta Pascali\inst{1}\orcidlink{0000-0001-7742-8126}, Sandro Sorbi\inst{4,6}\orcidlink{0000-0002-0380-6670} and
Paolo Matteini\inst{3}\orcidlink{0000-0002-8488-5867}} 

    \authorrunning{F. Conti et al.}
%
\institute{Institute of Information Science and Technologies ``A. Faedo'', \\ National Research Council of Italy,  Pisa (IT)\\ \email{\{Name.Surname\}@isti.cnr.it}
\and Department of Mathematics, University of Pisa,  Pisa (IT)
\and Institute of Applied Physics ``N. Carrara'', National Research Council of Italy, Sesto Fiorentino (IT)
\and Department of Neuroscience, Psychology, Drug Research and Child Health, University of Florence, Florence (IT)
\and Department of Clinical and Experimental Biomedical Sciences ``Mario Serio'', University of Florence, Florence (IT)
\and IRCCS Fondazione Don Carlo Gnocchi, Florence (IT)}

\maketitle              
\begin{abstract}
The cerebrospinal fluid (CSF) of  19 subjects who received a clinical diagnosis of Alzheimer's disease (AD) as well as of 5 pathological controls have been collected and analysed by Raman spectroscopy (RS). We investigated whether the raw and preprocessed Raman spectra could be used to distinguish AD from controls. First, we applied standard Machine Learning (ML) methods obtaining unsatisfactory results. Then, we applied ML to a set of topological descriptors extracted from raw spectra, achieving a very good classification accuracy ($>87\%$). 
Although our results are preliminary, they indicate that RS and topological analysis together may provide an effective combination to confirm or disprove a clinical diagnosis of AD. The next steps will include enlarging the dataset of CSF samples to validate the proposed method better and, possibly, to understand if topological data analysis could support the characterization of AD subtypes.


\end{abstract}

\section{Introduction}
Alzheimer’s disease (AD) affects tens of millions of people worldwide, being the most common neurodegenerative disease. Due to the population aging, the number of people affected by AD and other forms of dementia is expected to reach about 152 million by 2050 (World Alzheimer Report 2021 provided by Alzheimer’s Disease International, McGill University \url{https://www.alzint.org/resource/world-alzheimer-report-2021/}). 
At present, the clinical diagnosis of AD requires a series of neurological examinations (National Institute of Aging – Alzheimer’s Association criteria), while the definitive diagnosis is possible only after the patient’s death and brain tissue analysis. Therefore, there is a need to improve the accuracy of clinical diagnosis with innovative, cost-effective and specific approaches. Raman spectroscopy (RS) represents a fast, efficient, non-invasive diagnostic tool \cite{Eberhardt2015}, and the high-precision detection of RS is expected to reduce or replace other AD diagnostic tests. Recently, Raman-based techniques demonstrated significant potential in identifying AD by detecting specific biomarkers in body fluids \cite{polikretis2022}. Given the increasing number of RS studies, a systematic evaluation of the accuracy of RS in the diagnosis of AD was already performed, showing that RS is an effective and accurate tool for diagnosing AD, though it still cannot rule out the possibility of misdiagnosis \cite{Xu2023}. Recently, Raman spectroscopy of tissue samples has been coupled with topological machine learning to support the grading of bone cancer \cite{2023Conti}, showing the feasibility of a topological approach for multi-label classification.
The detection of CSF biomarkers is one of the diagnostic criteria for AD \cite{Blennow2018} because CSF is more sensitive than blood or other biofluids in the diagnosis of AD. Therefore, RS can be used as an effective tool to analyze CSF samples, as shown previously \cite{RYZHIKOVA2021,Huang2017}.

Here we propose a novel method based on the collection of the vibrational Raman fingerprint of the proteomic content of cerebrospinal fluid (CSF) and on the topological machine learning analysis of the Raman spectra in order to support the AD diagnosis. The achieved results encourage to keep on investigating topological machine learning tools, not only to establish more safely the proposed methodology by enlarging the experimentation but also to understand if looking at Raman spectra of CSF with the topological lens could also help to characterize AD subtypes.

\section{Population study and Data Acquisition}
The study population is made of 24 patients, enrolled in the framework of the Bando Salute 2018 PRAMA project (``Proteomics, RAdiomics \& Machine learning-integrated strategy for precision medicine for Alzheimer’s''), co-funded by the Tuscany Region, with the approval of the Institutional Ethics Committee of the Careggi University Hospital Area Vasta Centro (ref. number 17918\_bio). All of them showed pathological symptoms: the majority of them, 19 subjects, have been diagnosed with AD, while the others have been considered as controls (noAD), even if diagnosed with other neurological conditions: one with vascular dementia, three with hydrocephalus and one with multiple sclerosis. 

The CSF samples were collected by lumbar puncture, then immediately centrifuged at $200 g$ for 1 min, $20$ °C and stored at $-80$ °C until analysis \cite{Tashjian2019,CSFpreproc2012}. On the day of analysis, CSF samples were thawed and centrifuged again at $4000 g$ for 10 min at $4$ °C. The pellet was separated from the supernatant and further used for the analyses. A $2$ ul drop of the pellet was deposited onto a gold mirror support (ME1S-M01; Thorlabs, Inc., Newton, NJ), followed by air drying for 30 minutes and acquisition of Raman spectra from the outer ring of the dried drop. A set of five Raman spectra have been collected for each drop-casted sample by using a micro-Raman spectrometer (Horiba, France) in back-scattering configuration, equipped with a laser excitation source tuned at $785$ nm ($40$ mW power, $20$ second integration time, $10$ accumulations) and a Peltier cooled CCD detector.  

Finally, a set of five Raman spectra have been collected for each biological sample. In some cases, the same procedure has been replicated two or three times; it resulted in a dataset of 30 acquisitions of RS: 22 belonging to the AD class and 8 to the noAD class. 

\section{Methods}







After the Raman spectra are acquired, the data enter the following pipeline to return the final predictive model with classification accuracy. For each patient, the average of the five acquisitions of the raw Raman spectrum is computed. Next, the following transformations are applied to the RS: Fourier transform, Welch transform and autocorrelation. We applied the pipeline individually on the original spectra and on each of the transformations listed above. These computations were performed using the Python package SciPy~\cite{2020SciPy-NMeth}.

The spectra enter the pipeline of Topological Machine Learning (TML). For more detailed information on the pipeline, refer to~\cite{math10173086}. The pipeline performs a lower star filtration to extract the Persistence Diagrams (PDs). Since the data is a 1D spectrum, the only non-trivial homology group is $H_0$. The PD is vectorized using the following vectorization methods: Persistence Image~\cite{adams2017persistence} with parameters $\sigma \in \{0. 1, 1, 10\},\,\, n \in \{5, 10, 25\}$, Persistence Landscape~\cite{bubenik2015statistical}, Persistence Silhouette~\cite{chazal2014stochastic} and Betti curve~\cite{umeda2017time} all with parameters $n \in \{25, 50, 75, 100\}$. Finally, these vectors enter one of the following Machine Learning (ML) classifiers: Support Vector Classifier~\cite{cortes1995support}, Random Forest Classifier~\cite{breiman2001random} and Ridge Classifier~\cite{hoerl1970ridge}. The validation scheme of the pipeline is the Leave One Patient Out cross-validation (LOPO). This scheme is a generalization of the classic leave one out cross-validation~\cite{hastie2009elements}, with the difference that all data from the same patient are recursively left in the validation set, instead of a single data. This avoids biased high accuracy due to the similarity of the data coming from the same patient that may otherwise be found both in the training and validation set.

\begin{figure}[h]
\centering
\includegraphics[width=9.5cm]{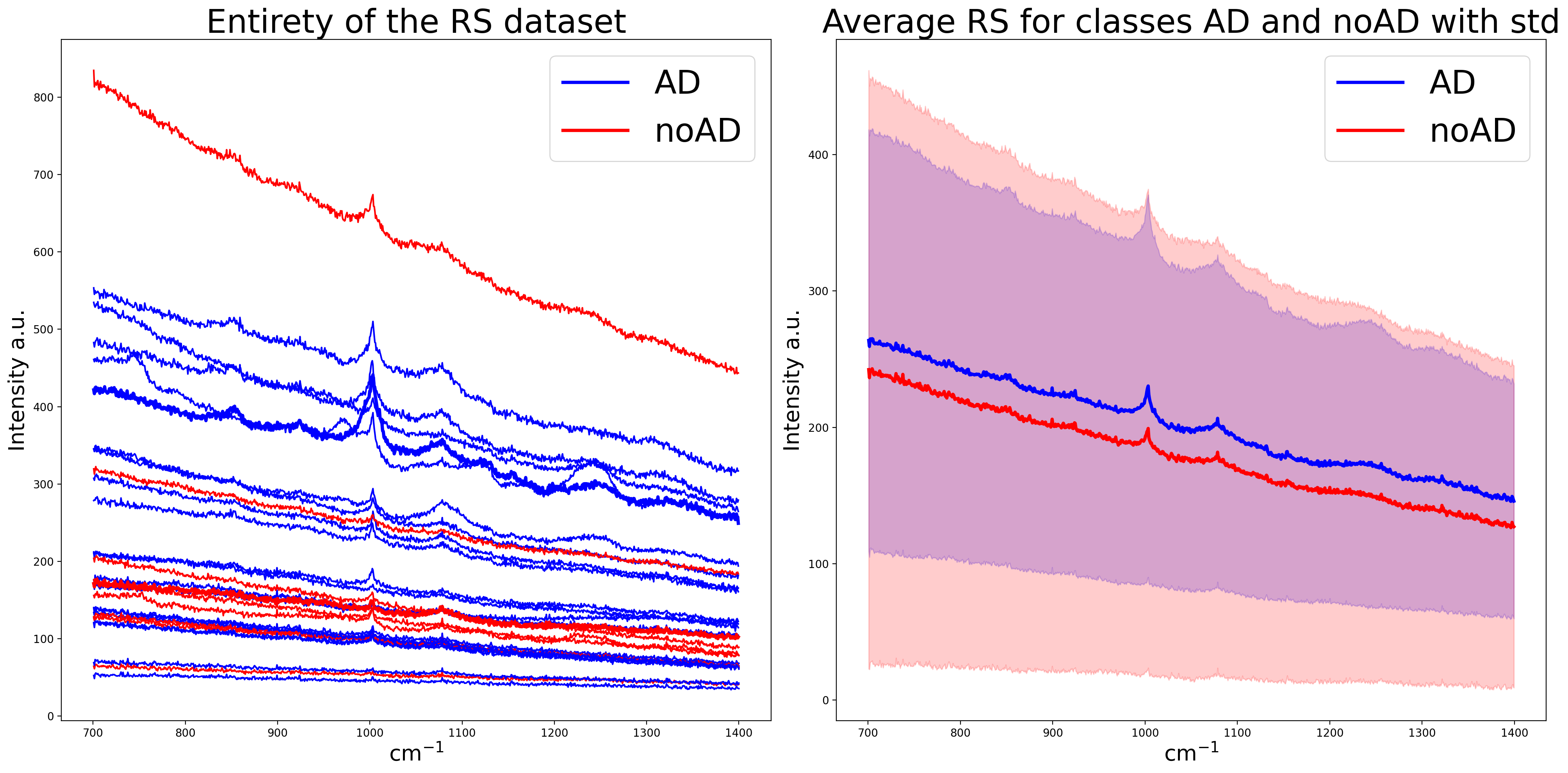}
\caption{(\textbf{a}) The entirety of the dataset of Raman spectra coloured by respective class. (\textbf{b}) The corresponding average with standard deviation.\label{fig:AllRS}}
\end{figure}

In Figure~\ref{fig:AllRS}, we report the entirety of the dataset of Raman spectra, divided by the class of Alzheimer's disease and the corresponding average with standard deviation. 

\begin{figure}[h]

\centering
\includegraphics[width=9.5cm]{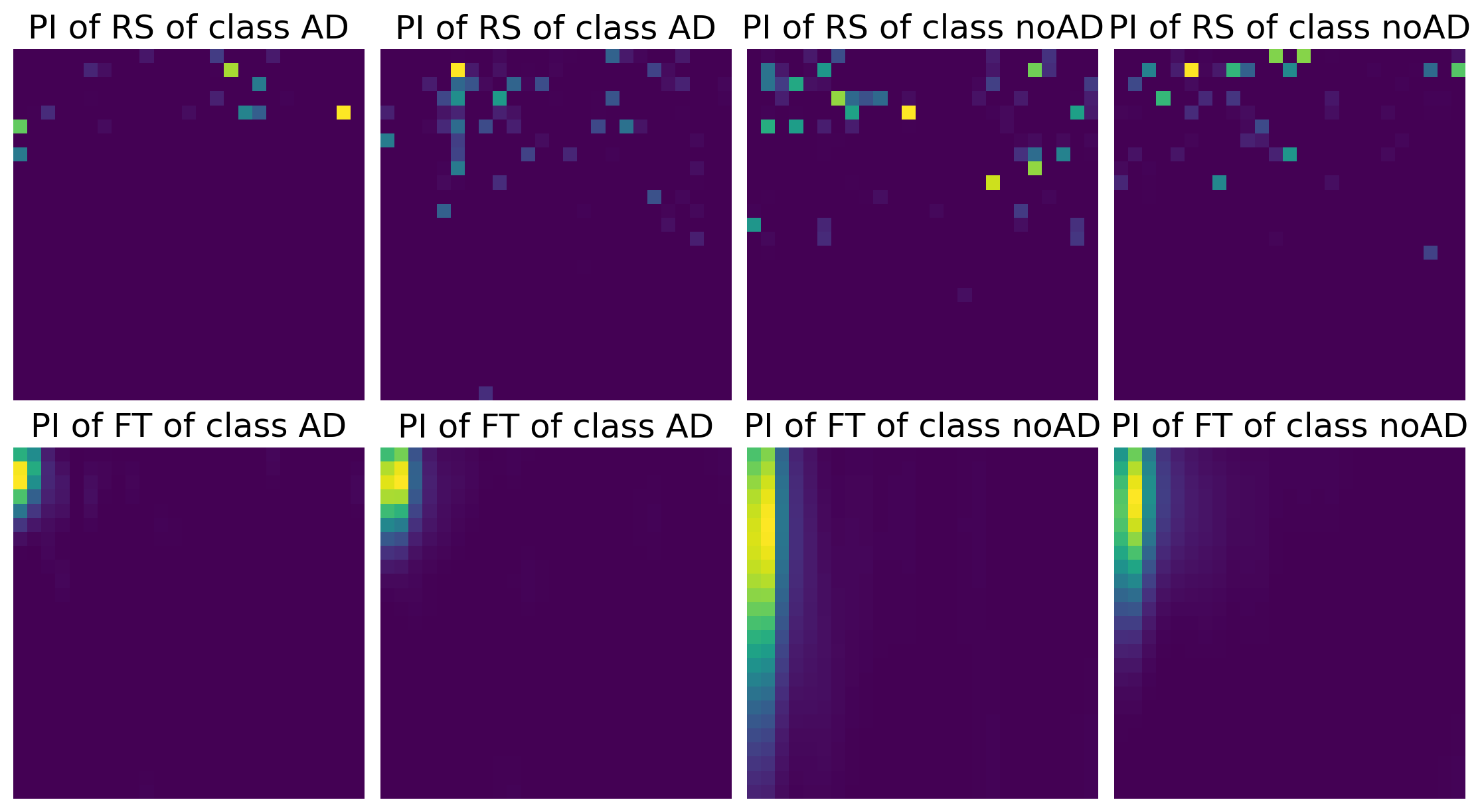}

\caption{First row: Two persistence images (PI) of the Raman spectra for class AD and two for class noAD. Second row: two PI of the Fourier transform for class AD and for class noAD. It appears that the PI obtained from Raman spectra are more chaotic between the two classes. In the PI obtained from the Fourier transform, a clearer division between AD and noAD is observed, with the latter having a more elongated dot.\label{fig:PI}}
\end{figure}

\begin{figure}[h]

\centering
\includegraphics[width=9.5cm]{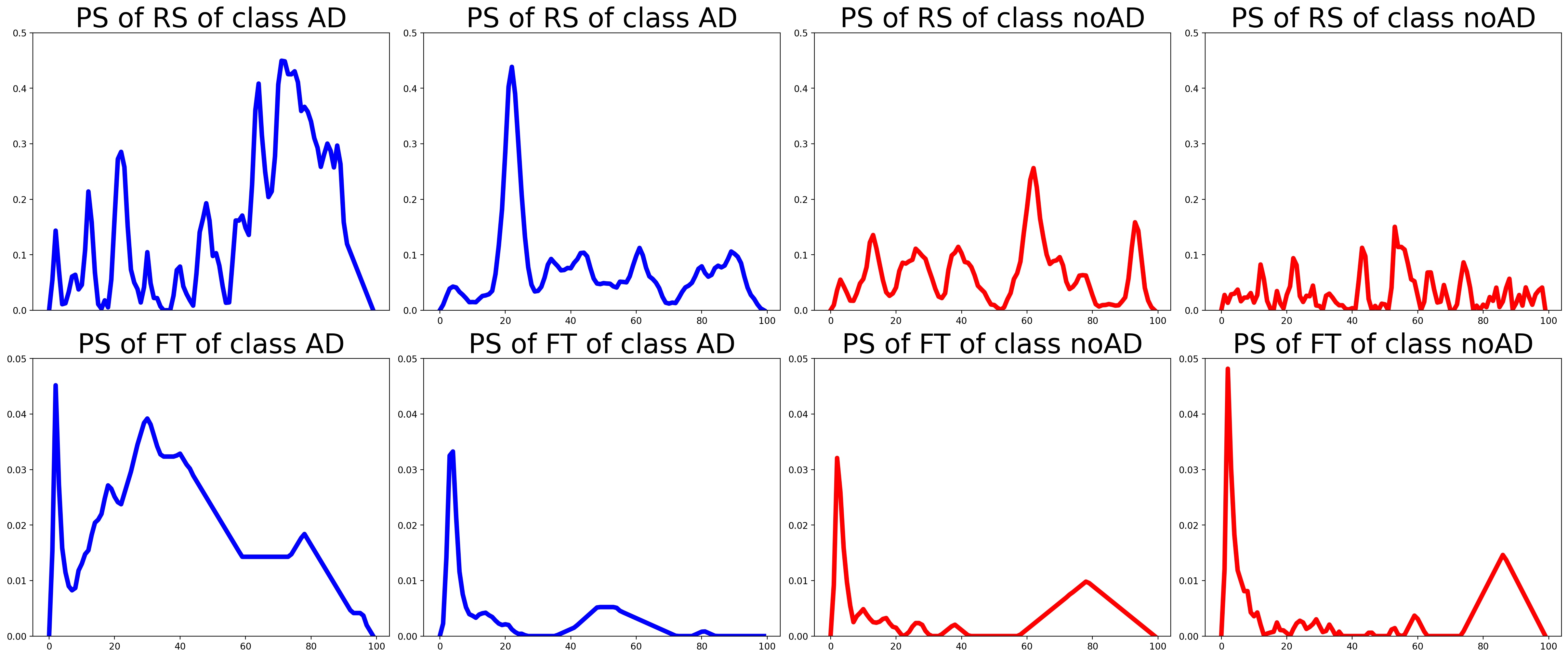}

\caption{First row: Two persistence silhouettes (PS) of the Raman spectra for class AD and for class noAD. Second row: two PS of the Fourier transform for class AD and for class noAD. Again, it seems that the PS coming from the RS are more chaotic, while in the PS coming from the Fourier transform there is a clearer division between AD and noAD, with the latter having a clear peak at the tail end of the signal.\label{fig:PS}}
\end{figure}

In Figure~\ref{fig:PI} are shown eight Persistence images, two for each combination AD-noAD and RS-FT. It is interesting to note that in the PIs coming from the Raman spectra the pattern seems more chaotic between the two classes, while in the PIs coming from the Fourier transform there is a clearer division. More in detail, the lit pixels in the PIs of class noAD have a more elongated shape than those of the AD class. This corroborates the results achieved in Section~\ref{sec:result}. In Figure~\ref{fig:PS} are shown eight Persistence silhouettes in the same fashion as Figure~\ref{fig:PI}. Again, there is a clearer division for the PSs coming from the Fourie transform. A peak at the tail end of the signal is present for the noAD class.

\section{Results}
\label{sec:result}

The results obtained from the pipeline on each of the transformations are shown in Table~\ref{tab:res}. The Fourier transform is the one that gets clearly the better results. We used as baseline accuracy the value of $0.733$, due to the imbalance in classes (i.e. the accuracy achieved by the classifier assigning always the most frequent label to any sample). 

\begin{table}[h] 
\caption{Accuracy result of the TML pipeline for different input.\label{tab:res}}
\newcolumntype{C}{>{\centering\arraybackslash}X}
\begin{tabularx}{\textwidth}{CCC}
\toprule
\textbf{Method}	    & \textbf{Accuracy}	& \textbf{Vectorization and Classifier}\\
\midrule
$H_0$		        & $0.833$			& PI and Ridge\\
Fourier transform	& $0.875$			& PS and SVC\\
Welch transform		& $0.763$			& PI and SVC\\
Autocorrelation		& $0.667$			& PI and Ridge\\
\bottomrule
\end{tabularx}
\end{table}

It is worth pointing out that even standard preprocessing applied to Raman spectra could lead to a classification accuracy below the baseline accuracy. This is probably due to the fact that in our dataset, the signal-to-noise ratio is quite low. On the other hand, the accuracy value ($> 83\%$) achieved by extracting $H_0$ features from raw spectra is in line with results of \cite{RYZHIKOVA2021}, while results achieved by extracting topological features after performing the Fourier transform are even better ($87.5\%$). 

\section{Discussion}
The results described above support strongly that RS and topological analysis together may provide an effective combination to confirm or disprove a clinical diagnosis of AD. Also, the training of the classification ML model trained on the topological features extracted from the Raman spectra acquired on CSF sample does not need the choice or set of any parameters; hence, the proposed methodology may evolve in automatic support to AD diagnosis, which could be easily embedded in a commercial platform of Raman spectroscopy. 
The above considerations are preliminary and require further confirmation from the statistical viewpoint. From this perspective, the next steps will include enlarging the dataset of CSF samples to validate the proposed method better and, possibly, to understand if topological machine learning could support the characterization of AD subtypes.

%
%
%
%

\bibliographystyle{splncs04}

\bibliography{bibliovision}






\end{document}